\documentclass[conference]{IEEEtran}
\IEEEoverridecommandlockouts

\usepackage{amsmath,amssymb,amsfonts}
\usepackage{algorithmic}
\usepackage{graphicx}
\usepackage{textcomp}
\usepackage{xcolor}
\usepackage{url}
\usepackage[hidelinks]{hyperref}
\def\BibTeX{{\rm B\kern-.05em{\sc i\kern-.025em b}\kern-.08em
    T\kern-.1667em\lower.7ex\hbox{E}\kern-.125emX}}
\begin{document}

\title{Federated Learning for Video Violence Detection: Complementary Roles of Lightweight CNNs and Vision-Language Models for Energy-Efficient Use}

\author{\IEEEauthorblockN{Sébastien Thuau}
\IEEEauthorblockA{\textit{esieaLab, ETIS Laboratory} \\
\textit{ESIEA, University of CY Cergy}\\
Paris, France \\
thuau@et.esiea.fr}
\and
\IEEEauthorblockN{Siba Haidar}
\IEEEauthorblockA{\textit{esieaLab} \\
\textit{ESIEA}\\
Paris, France \\
siba.haidar@esiea.fr}
\and
\IEEEauthorblockN{Rachid Chelouah}
\IEEEauthorblockA{\textit{ETIS Laboratory, CNR1S, UMR8051} \\
\textit{University of CY Cergy}\\
Paris, France \\
rc@cy-tech.fr}
}

\maketitle

\begin{abstract}
Deep learning–based video surveillance increasingly demands privacy‐preserving architectures with low computational and environmental overhead. Federated learning preserves privacy but deploying large Vision-Language Models (VLMs) introduces major energy and sustainability challenges.
We compare three strategies for federated violence detection under realistic non-IID splits on RWF-2000 and RLVS datasets: zero-shot pretrained VLM inference, LoRA-based fine-tuning of LLaVA-NeXT-Video-7B, and personalized federated learning of a 65.8 M-parameter 3D CNN.
All methods exceed 90\% accuracy in binary violence detection. The 3D CNN achieves superior calibration (ROC AUC 92.59\%) at roughly half the energy cost (240 Wh vs.~570 Wh) of federated LoRA, while VLMs provide richer multimodal reasoning. Hierarchical category grouping—based on semantic similarity and class exclusion—boosts VLM multiclass accuracy from 65.31\% to 81\% on the UCF-Crime dataset.

To our knowledge, this is the first comparative simulation study of LoRA-tuned VLMs and personalized CNNs for federated violence detection, with explicit energy and CO$_2$e quantification. Our results inform hybrid deployment strategies that default to efficient CNNs for routine inference and selectively engage VLMs for complex contextual reasoning.
\end{abstract}


\section{Introduction}

Deep learning–based video surveillance systems are increasingly deployed to detect and analyze violent incidents in public spaces, yet centralized training introduces privacy risks by transferring sensitive footage off‐device and incurs substantial computational and environmental overhead.

Vision-Language Models (VLMs) enhance multimodal reasoning through natural language interfaces but carry billions of parameters, challenging energy efficiency and carbon footprint constraints in federated learning environments, which—while preserving privacy—do not inherently address the resource demands of such architectures.

Aligned with the goals of the DIVA (Decentralized Intelligence for Visual Awareness) initiative, we investigate whether large‐scale VLMs can be fine‐tuned sustainably under non-IID federated conditions for violence detection and compare their energy consumption and emissions against lightweight 3D CNNs with competitive performance.

\paragraph*{Model naming convention.}
After first mention, we use the following abbreviations: \emph{LLaVA-NeXT-Video-7B} (\textbf{LLaVA}); \emph{Qwen2.5-VL-7B-Instruct} (\textbf{QwenVL}); \emph{Qwen2.5-Omni-7B} (\textbf{QwenOmni}); \emph{Ovis2 (8B)} (\textbf{Ovis-8B}); \emph{Ovis2 (4B)} (\textbf{Ovis-4B}); \emph{InternVL3-8B} (\textbf{InternVL-8B}); \emph{MiniCPM-o-2.6} (\textbf{MiniCPM}). We also refer to \emph{Personalized Federated Learning} as (\textbf{PFL}) and to the personalized 3D convolutional network as (\textbf{3D CNN}).

To this end, we assess zero‐shot inference of a VLM baseline, federated LoRA fine‐tuning of LLaVA, and personalized federated learning of a compact 3D CNN for binary classification on RWF‐2000 and RLVS datasets with realistic non-IID splits. Our key contributions include the first comparative study of LoRA‐tuned VLMs versus personalized CNNs for federated violence detection, a systematic quantification of energy use, communication overhead, and CO$_2$e emissions, evaluation under distributed non-IID conditions, and practical insights for hybrid deployment strategies that balance accuracy, privacy, and sustainability.

\section{Related Work}

\subsection{Privacy-Preserving Architectures for Video Surveillance and Energy-Efficient Multimodal AI Models}

Federated learning (FL) mitigates privacy risks inherent to centralized video analytics by keeping raw footage on-device and exchanging only model updates or task-specific features. Compared to server-side ingestion of video streams, FL removes large-scale data transfers and reduces exposure of personally identifiable information. Prior systems for video-based distributed surveillance train local models at the edge and synchronize via secure aggregation, preventing the server from inspecting client updates~\cite{jiang2022fedgnn,singh2022videoFL}. Surveyed frameworks further layer formal protections—differential privacy, homomorphic encryption, and multi-party computation—to strengthen confidentiality guarantees~\cite{abdulrahman2021survey,bellavista2021survey}. Complementary “semantic-first’’ pipelines discard raw frames after on-device processing and transmit only compact, AI-extracted representations, aligning with frugal deployment principles~\cite{vyas2024leaky}.

On the efficiency front, parameter-efficient fine-tuning (PEFT) enables low-cost personalization in FL. Low-Rank Adaptation (LoRA) factorizes weight updates into low-rank matrices, substantially reducing compute and communication. Recent federated LoRA frameworks report large efficiency gains: aggregation-aware and quantization-enhanced designs shrink update sizes with minor accuracy loss, while server-side correction mitigates aggregation bias without added bandwidth~\cite{ribeiro2024flocora,bian2024lora-fair}. For multimodal settings, FLORA applies rank-2 LoRA adapters to CLIP ViT-L/14 and demonstrates significant speedups and memory savings over full fine-tuning across diverse vision datasets~\cite{nguyen2024flora}. Additional variants address heterogeneity via shared homogeneous adapters, dual (global/personalized) adapters, and adaptive-rank tuning for constrained clients~\cite{yi2023fedlora,qi2024fdlora,su2024hafl}. Collectively, these methods fall into (i) aggregation-enhancing strategies (e.g., stacking or corrected aggregation) and (ii) adapter-level optimizations (quantization, personalization, adaptive rank).

Despite this progress, the literature seldom quantifies the cost–benefit trade-offs of federated fine-tuning for vision–language models (VLMs) versus lightweight CNNs under realistic non-IID surveillance data. Moreover, to our knowledge, no prior work evaluates LoRA-based federated VLM fine-tuning for violence detection under practical surveillance conditions or benchmarks its cost-effectiveness against lightweight CNNs under matched non-IID partitions—questions we address empirically.

\subsection{Environmental Impact Assessment}

At surveillance scale, the cumulative energy footprint of model training and adaptation across thousands of cameras is non-negligible, yet empirical transparency on consumption and emissions remains limited. Emerging policy and standardization efforts underscore the need for rigorous environmental accounting throughout the AI lifecycle. In the EU, the AI Act introduces obligations touching energy transparency and resource use for higher-risk systems, while French initiatives promote methodological baselines for “frugal’’ AI (e.g., AFNOR SPEC~2314) and institutional support via ADEME and related programs. Building on this context, we quantify energy and carbon for centralized vs.\ federated fine-tuning of VLMs, and compare against frugal CNN baselines under non-IID deployments. Our results provide evidence to guide sustainable surveillance model selection and adaptation.

\section{Methodology}
\subsection{Binary Classification}

\subsubsection{Experimental Design and Data Configuration}
We build a 4{,}000-clip corpus by concatenating RWF\textendash2000 and RLVS (2{,}000 samples each; balanced violence/non\textendash violence)~\cite{rwf2000,rlvs2020}. A stratified 80/20 split defines train/validation. To induce cross\textendash silo heterogeneity, data are partitioned over 10 clients by domain: clients 1–5 receive only RWF\textendash2000, clients 6–10 only RLVS. Additional non\textendash IID skew is introduced via Dirichlet sampling with concentration $\alpha{=}1$ at the client level.

\subsubsection{Distributed Learning Configuration}
All FL runs are simulated (no real devices). We train with FedAvg for 20 communication rounds~\cite{mcmahan2017communication}, emulating 10 clients (each: 1 GPU, 4 CPU cores). Each round, we sample 50\% of clients (5/10) to model partial participation due to bandwidth/availability.

\subsubsection{Vision\textendash Language Model Adaptation}
We apply LoRA~\cite{hu2021lora} to \textbf{LLaVA} (\textit{LLaVA\textendash NeXT\textendash Video\textendash7B})~\cite{zhang2024llavanextvideo,liu2024llavanext}. The visual encoder is CLIP ViT\textendash L/14~\cite{radford2021clip}, operating on 24\textendash frame clips. Prompts follow: \texttt{``Analyze the video. Is this a fight scene? Answer yes or no.''} Adapters use rank $r{=}16$, scaling $\alpha{=}32$, with 4\textendash bit quantization; only adapter weights are communicated. Adapter modules are inserted in attention blocks; base weights remain frozen.

\subsubsection{Comparative Baseline Architecture}
As a personalized baseline, we use a $\sim$65.8M\textendash param 3D\textendash CNN with parameter decoupling~\cite{tan2023towards}: a globally aggregated spatiotemporal backbone (shared features) and client\textendash local classification heads (specialization), enabling personalization while preserving cross\textendash domain transfer.

\subsubsection{Performance Assessment Protocol}
We evaluate under three regimes: (i) zero\textendash shot VLMs (\textbf{LLaVA}, \textbf{QwenVL}, \textbf{QwenOmni}, \textbf{Ovis-8B}, \textbf{InternVL-8B}, \textbf{MiniCPM}), (ii) federated LoRA fine\textendash tuning, and (iii) personalized CNN training. Metrics: accuracy and F1 Score (reported as \%). Environmental impact is computed with CodeCarbon~\cite{codecarbon}, converting metered energy (kWh) to \mbox{gCO$_2$e} using a Paris\textendash region factor of 56~\mbox{gCO$_2$e}/kWh (assumption). Hardware: NVIDIA A10; results are averaged over three seeds.

\subsubsection{Theoretical Basis for Environmental Impact Modeling}

We propose a dual framework for modeling the carbon footprint of machine learning workflows—distinguishing runtime emissions from hardware-related manufacturing costs—offering a scalable foundation when empirical measurement is limited or impractical.

\subsection{Multiclass Classification}

To probe beyond binary decisions, we evaluate VLM comprehension on multiclass anomaly recognition using UCF\textendash Crime, a large\textendash scale surveillance benchmark covering 13 anomaly categories \emph{plus a Normal class}~\cite{sultani2019realworld,crcv2019realworld}. As a zero\textendash shot baseline, we run inference on 20\% of the data with three VLMs reused from the binary study (\textbf{QwenVL}, \textbf{Ovis-8B}, \textbf{LLaVA}), then extend inference to the full dataset to validate error patterns. Confusion analysis reveals systematic overlaps among semantically close classes. We therefore adopt a hierarchical grouping on the full set to reduce ambiguity while preserving operational relevance:\label{refinedClasses}
\begin{itemize}
    \item \textbf{Destruction}: Arson, Explosion
    \item \textbf{Property Crime}: Burglary, Robbery, Shoplifting, Stealing, Vandalism
    \item \textbf{Violence}: Assault, Fighting
\end{itemize}
\textit{RoadAccidents} and \textit{Normal Video} are retained as distinct categories; \textit{Abuse}, \textit{Arrest}, and \textit{Shooting} are excluded. The inclusion/exclusion rationale and downstream effects on accuracy and calibration are detailed in Section IV.

\section{Experiments and Results}
\subsection{Binary Classification}

\subsubsection{Exp. 1.1: Zero-Shot VLM Inference: Accuracy vs. Efficiency}

We assessed the performance–efficiency trade-offs of pretrained vision-language models on binary violence classification without fine-tuning.
For readability and consistency with the above convention, we use \textbf{LLaVA}, \textbf{QwenVL}, \textbf{QwenOmni}, \textbf{Ovis-8B}, \textbf{InternVL-8B}, and \textbf{MiniCPM} throughout the rest of the paper. Performance evaluation on RLVS (2,000 videos) revealed that \textbf{Ovis-8B}~\cite{lu2024ovis} achieved optimal accuracy (95.80\%) and F1 Score (96.0\%) while maintaining reasonable energy consumption (457 Wh). \textbf{LLaVA}~\cite{zhang2024llavanextvideo} demonstrated competitive performance (94.15\% accuracy) with lower energy requirements (328 Wh), as shown in Table~\ref{tab:vlm-rlvs-performance}.

\begin{table}[ht]
\caption{Zero-shot VLM performance on RLVS dataset (2000 videos), mean ± std over 3 runs}
\centering
\scriptsize
\begin{tabular}{|l|c|c|c|}
\hline
\textbf{Model} & \textbf{Acc. (\%)} & \textbf{F1 Score(\%)} & \textbf{Energy (Wh)}  \\
\hline
\textbf{Ovis-8B~\cite{lu2024ovis}}   & \textbf{95.80 ± 0.15} & \textbf{96.0 ± 0.4} & \textbf{457 ± 7}\\
LLaVA~\cite{zhang2024llavanextvideo} & 94.15 ± 0.20 & 94.0 ± 5.0 & 328 ± 6\\
QwenVL~\cite{qwen2.5-VL} & 91.00 ± 0.25 & 91.0 ± 6.0 & 616 ± 8\\
InternVL-8B~\cite{InternVL3-8B} & 92.75 ± 0.22 & 92.0 ± 5.0 & – \\
MiniCPM~\cite{minicpm} & 87.75 ± 0.30 & 86.0 ± 6.0 & – \\
Ovis-4B~\cite{lu2024ovis} & 60.75 ± 0.45 & – & – \\
\hline
\multicolumn{4}{l}{\parbox[t]{0.9\linewidth}{– indicates missing values. Energy in Wh. \\Emissions: 56 g CO$_2$e/kWh = 0.056 g/Wh.}}
\end{tabular}
\vspace{1mm}
\label{tab:vlm-rlvs-performance}
\end{table}

A comprehensive benchmark on 800 videos (400 RLVS + 400 RWF-2000) confirmed model performance variations across architectures. Table~\ref{tab:comprehensive-benchmark} demonstrates \textbf{Ovis-8B}'s superior performance (92.5\% accuracy, 92.5\% ROC AUC) with efficient resource utilization (172 Wh, 1928 s inference time).

\begin{table}[ht]
\caption{Comprehensive zero-shot evaluation on 800 surveillance videos (mean ± std over 3 runs).}
\centering
\scriptsize
\resizebox{\linewidth}{!}{%
\begin{tabular}{|l|c|c|c|c|c|}
\hline
\textbf{Model} & 
\begin{tabular}{@{}c@{}}\textbf{Accuracy} \\ (\%)\end{tabular} & 
\begin{tabular}{@{}c@{}}\textbf{F1 Score} \\ (\%)\end{tabular} & 
\begin{tabular}{@{}c@{}}\textbf{ROC AUC} \\ (\%)\end{tabular} & 
\begin{tabular}{@{}c@{}}\textbf{Energy} \\ (Wh)\end{tabular} & 
\begin{tabular}{@{}c@{}}\textbf{Duration} \\ (s)\end{tabular} \\
\hline
Ovis-8B & 92.5 ± 0.2  & 92.5 ± 0.4  & 92.5 ± 0.6  & 172 ± 4  & 1928 ± 30 \\
LLaVA & 90.4 ± 0.3  & 90.6 ± 0.5  & 86.4 ± 0.7  & 200 ± 5  & 3215 ± 46 \\
InternVL-8B & 88.5 ± 0.3  & 88.3 ± 0.4  & 88.5 ± 0.6  & 130 ± 4  & 1745 ± 35 \\
QwenVL & 88.5 ± 0.3  & 87.4 ± 0.5  & 88.5 ± 0.5  & 660 ± 12 & 5770 ± 55 \\
\hline
\multicolumn{6}{l}{\parbox[t]{1.1\linewidth}{ROC AUC scores were computed only when the model produced consistent confidence outputs across classes; otherwise, metrics were omitted or approximated from logits or ranking confidence.}}
\end{tabular}
}
\label{tab:comprehensive-benchmark}
\end{table}

Carbon emissions were estimated using regional grid intensity (56 g CO$_2$e/kWh) with CodeCarbon~\cite{codecarbon} monitoring. Results demonstrated significant variance in energy efficiency across architectures, with LLaVA selected for subsequent federated experiments due to its balanced performance–efficiency profile and LoRA compatibility.

\subsubsection{Exp. 1.2: Federated Fine-Tuning of LLaVA with LoRA}

We implemented parameter-efficient federated adaptation using LoRA~\cite{hu2021lora} on LLaVA under non-IID conditions. The experimental setup partitioned 4,000 videos across 10 clients with domain-based heterogeneity: clients 1-5 received RWF-2000 content, clients 6-10 processed RLVS data. Label imbalance was induced by sampling from a Dirichlet distribution with a concentration parameter of one. All federated experiments employed partial participation, with 5 out of 10 clients selected at random in each round to reflect realistic connectivity and bandwidth limitations.

The federated optimization employed FedAvg~\cite{mcmahan2017communication} with partial participation (5/10 clients per round) over 20 communication epochs. LoRA adapters utilized rank-16 configuration with scaling factor $\alpha=32$, implemented through 4-bit quantization via the Flower federated learning framework~\cite{beutel2020flower}. Differential privacy mechanisms were integrated using fixed clipping (norm=1) to ensure convergence stability.

Table~\ref{tab:federated-lora-results} presents the comparative analysis between federated LoRA fine-tuning and zero-shot baseline. Federated adaptation significantly improved model calibration (log loss: 0.706 → 0.535) and discriminative performance (ROC AUC: 85.93\% → 91.24\%) while maintaining competitive accuracy.

\begin{table}[ht]
\caption{Federated LoRA adaptation vs. zero-shot baseline on 800-video validation set}
\centering
\scriptsize
\begin{tabular}{|p{3.3cm}|c|c|}
\hline
\textbf{Metric} & \textbf{LoRA FL} & \textbf{Zero-Shot} \\
\hline
Training Duration (s)     & 4741     & –        \\
Energy (Wh)               & 570      & 200      \\
Emissions (g CO$_2$e)     & 32.0     & 11.2      \\
Accuracy (\%)             & 90.87    & 90.62    \\
F1 Score (\%)             & 90.93    & 90.84    \\
ROC AUC  (\%)             & 91.24    & 85.93    \\
Log loss                  & 0.535    & 0.706    \\
\hline
\multicolumn{3}{l}{\parbox[t]{0.9\linewidth}{Validation: 800 videos (20\% of RWF + RLVS). Mean over 3 runs. FL = Federated Learning.}}
\end{tabular}
\vspace{1mm}
\label{tab:federated-lora-results}
\end{table}

Training consumed 570 Wh over 4,741 seconds, generating 32.0 g CO$_2$e emissions. The efficiency gains from transmitting only 4-bit adapter weights demonstrated the viability of communication-efficient federated adaptation for resource-constrained surveillance environments.

\subsubsection{Exp. 1.3: Personalized Federated Learning with Lightweight 3D CNN}

We evaluated a parameter-decoupled 3D convolutional architecture (65.8M parameters) using personalized federated learning (PFL)~\cite{tan2023towards}. The model employed shared spatiotemporal feature extraction layers (globally aggregated) and client-specific classification heads (locally optimized), balancing generalization with domain adaptation.

The federated configuration mirrored previous experiments: 10 clients with domain-based partitioning, 20 communication rounds, and 50\% participation rate. 
Table~\ref{tab:architecture-comparison} presents comparative performance across the three evaluated approaches.

\begin{table}[ht]
\caption{Performance comparison: 3D CNN vs. LLaVA variants (20 rounds, 5 clients/round)}
\centering
\scriptsize
\setlength{\tabcolsep}{3pt}
\begin{tabular}{|l|c|c|c|}
\hline
\textbf{Metric} & \textbf{3D CNN PFL} & \textbf{LLaVA ZS} & \textbf{LLaVA LoRA} \\
\hline
Train Time (s) & 3795 & – & 4741 \\
Energy (Wh) & 240 & 200 & 570 \\
CO$_2$e (g) & 13.4 & 11.2 & 31.9 \\
Acc. (\%) & 90.75 & 90.62 & 90.87 \\
F1 Score(\%) & 90.66 & 90.84 & 90.93 \\
AUC (\%) & \textbf{92.59} & 85.93 & 91.24 \\
Log loss & \textbf{0.546} & 0.707 & 0.535 \\
\hline
\multicolumn{4}{l}{\parbox[t]{.9\linewidth}{ZS = zero-shot, PFL = Personalized FL. 3D CNN and LoRA: non-IID validation (800 videos). ZS: balanced centralized test set (800 videos).}}
\end{tabular}
\vspace{1mm}
\label{tab:architecture-comparison}
\end{table}

Energy efficiency analysis demonstrated superior resource utilization: 240 Wh consumption and 13.4 g CO$_2$e emissions during complete federated training. The 3D CNN achieved the highest ROC AUC (92.59\%) and the lowest log loss (0.546) among evaluated approaches.

\subsection{Multiclass Classification with Vision-Language Models}

\subsubsection{Zero-shot Baseline Evaluation}

To evaluate the performance of vision-language models (VLMs) on multiclass crime detection, we conducted inference on the complete UCF-Crime dataset using zero-shot classification. This approach allows us to assess the models' inherent understanding of criminal activities without task-specific fine-tuning.

\paragraph{Results}

Table~\ref{tab:accuracy-f1-comparison} presents the overall performance metrics for all three models on the whole dataset of UCF-Crime, with the initial 14 classes as defined in the dataset.

\begin{table}[ht]
\caption{Zero-shot Baseline Evaluation Results (mean ± std over 3 runs)}
\centering
\scriptsize
\setlength{\tabcolsep}{6pt}
\begin{tabular}{|c|c|c|c|c|}
\hline
\textbf{Model} & \textbf{Accuracy (\%)} & \textbf{F1 Score (\%)} & \textbf{Energy (Wh)} & \textbf{CO$_2$e (g)} \\
\hline
\textbf{Ovis-8B} & 65.31 & 60.43 & 1149.8 & 64.4 \\
\hline
\textbf{LLaVA} & 47.77 & 38.01 & 385.2 & 21.6 \\
\hline
\textbf{QwenVL} & 51.26 & 48.35 & 597.6 & 33.5 \\
\hline
\end{tabular}
\label{tab:accuracy-f1-comparison}
\end{table}

These findings indicate marked disparities in model performance, with \textbf{Ovis-8B} surpassing all others on every metric. Accordingly, subsequent results are presented for \textbf{Ovis-8B}. With the best model achieving only 65.31\% accuracy, we investigated the factors underlying the VLMs’ difficulties.

\begin{figure}[htbp]
\centerline{\includegraphics[width=0.35\textwidth]{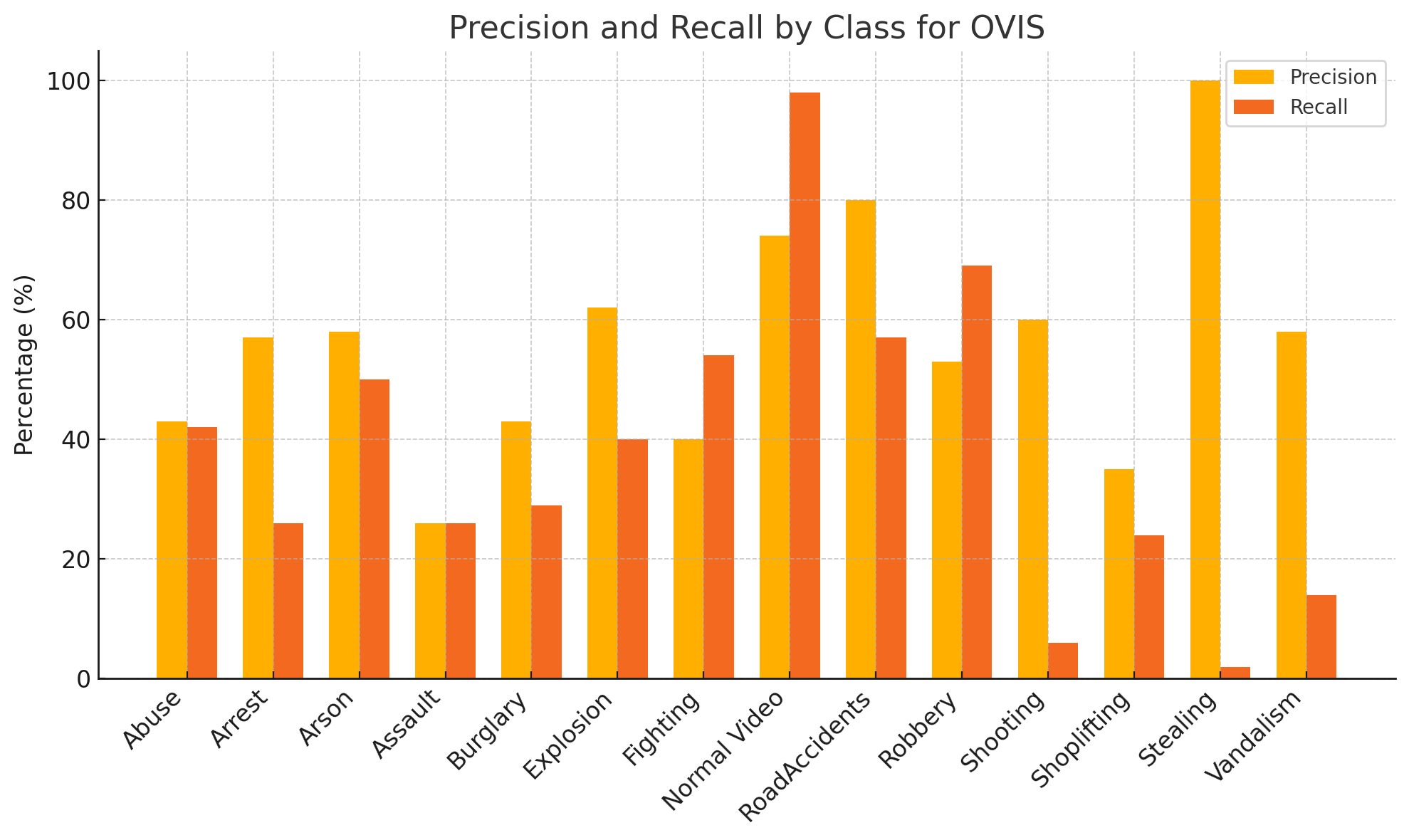}}
\caption{Precision and recall scores by class for Ovis-8B on the full UCF-Crime dataset. Significant variability across classes highlights the difficulty of fine-grained classification and supports the motivation for semantic grouping.}
\label{fig:precision_recall}
\end{figure}

\paragraph{Per-Class Analysis}
Figure~\ref{fig:precision_recall} illustrates the precision and recall performance for each category across \textbf{Ovis-8B}. The analysis reveals substantial variation in detection accuracy across different crime types, with certain categories proving more challenging for automated classification. Furthermore, the recall for “Normal Video” is exceptionally high because the models tend to label inputs as “Normal Video” when uncertain.

\begin{figure}[htbp]
\centerline{\includegraphics[width=0.5\textwidth]{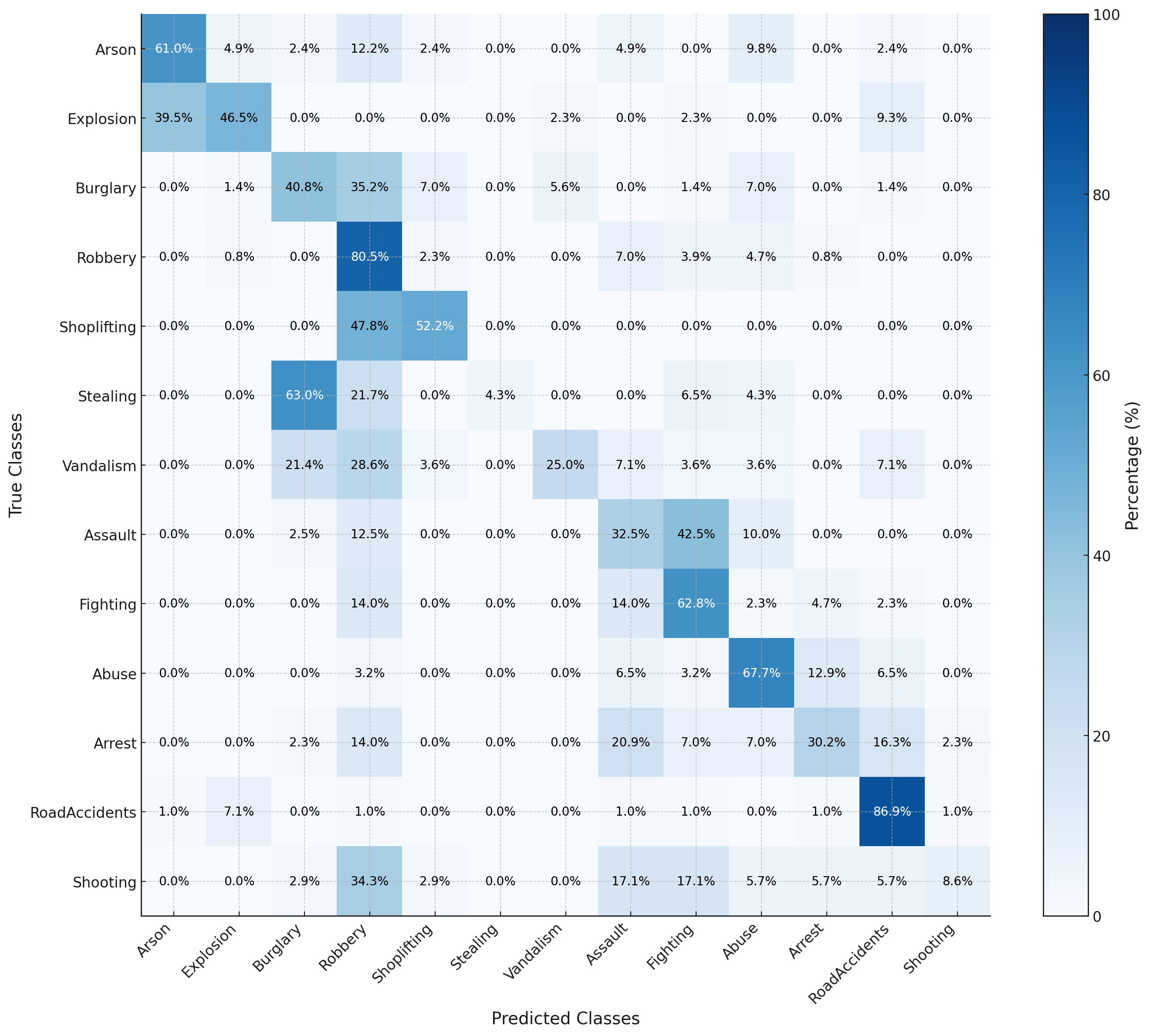}}
\caption{Normalized confusion matrix for zero-shot VLM predictions on the full UCF-Crime dataset, \emph{excluding} the ``Normal Video'' class. Highlights widespread misclassification between semantically adjacent classes and shows the tendency of VLMs to default to ``Normal'' when uncertain.}
\label{fig:confusion_analysis}
\end{figure}

Figure \ref{fig:confusion_analysis} shows the normalized confusion matrix with “Normal Video” excluded to focus on anomaly classification.

This confusion matrix (Fig.~\ref{fig:confusion_analysis}) highlights pronounced misclassifications among semantically adjacent categories by \textbf{Ovis-8B}. The three most significant confusions are:
\begin{enumerate}
  \item \textit{Arson} vs. \textit{Explosion}, since explosions frequently involve fire;
  \item \textit{Burglary}, \textit{Robbery}, \textit{Shoplifting}, \textit{Stealing}, and \textit{Vandalism}, all concerning offenses against property, where the model fails to distinguish subtle differences among property crimes;
  \item \textit{Assault} vs. \textit{Fighting}, both depicting interpersonal violence.
\end{enumerate}
Moreover, the \textit{Arrest} and \textit{Shooting} classes exhibit pervasive confusion with other anomalies and were therefore excluded from further experiments. The \textit{Abuse} class was excluded owing to the subtle and ambiguous nature of its instances.

\subsubsection{Refined Classification Strategy}

The revised classification strategy now focuses on aggregated categories as described in Section~\ref{refinedClasses}. We performed inference again on the full dataset, excluding the 150 videos from the omitted classes, and instructed the model to predict only the remaining categories. The same prompt structure was reused with grouped categories replacing the original 14 classes.

The results, shown in Table~\ref{tab:performance}, exhibit a marked improvement in performance, as expected given the reduced number of target classes.

\begin{table}[htbp]
\centering
\caption{Zero-shot Baseline with Refined Classification (mean ± std over 3 runs)}
\label{tab:performance}
\setlength{\tabcolsep}{6pt}
\begin{tabular}{|l|c|c|c|c|}
\hline
\textbf{Model} & \textbf{Accuracy (\%)} & \textbf{F1 Score (\%)} & \textbf{Energy (Wh)} & \textbf{CO$_2$e (g)} \\
\hline
Ovis-8B & 81.0 & 80.0 & 985.9 & 55.2 \\
\hline
LLaVA & 50.0 & 41.0 & 346.7 & 19.4 \\
\hline
QwenVL & 70.0 & 69.0 & 931.5 & 52.2 \\
\hline
\end{tabular}
\end{table}

\begin{figure}[htbp]
\centerline{\includegraphics[width=0.33\textwidth]{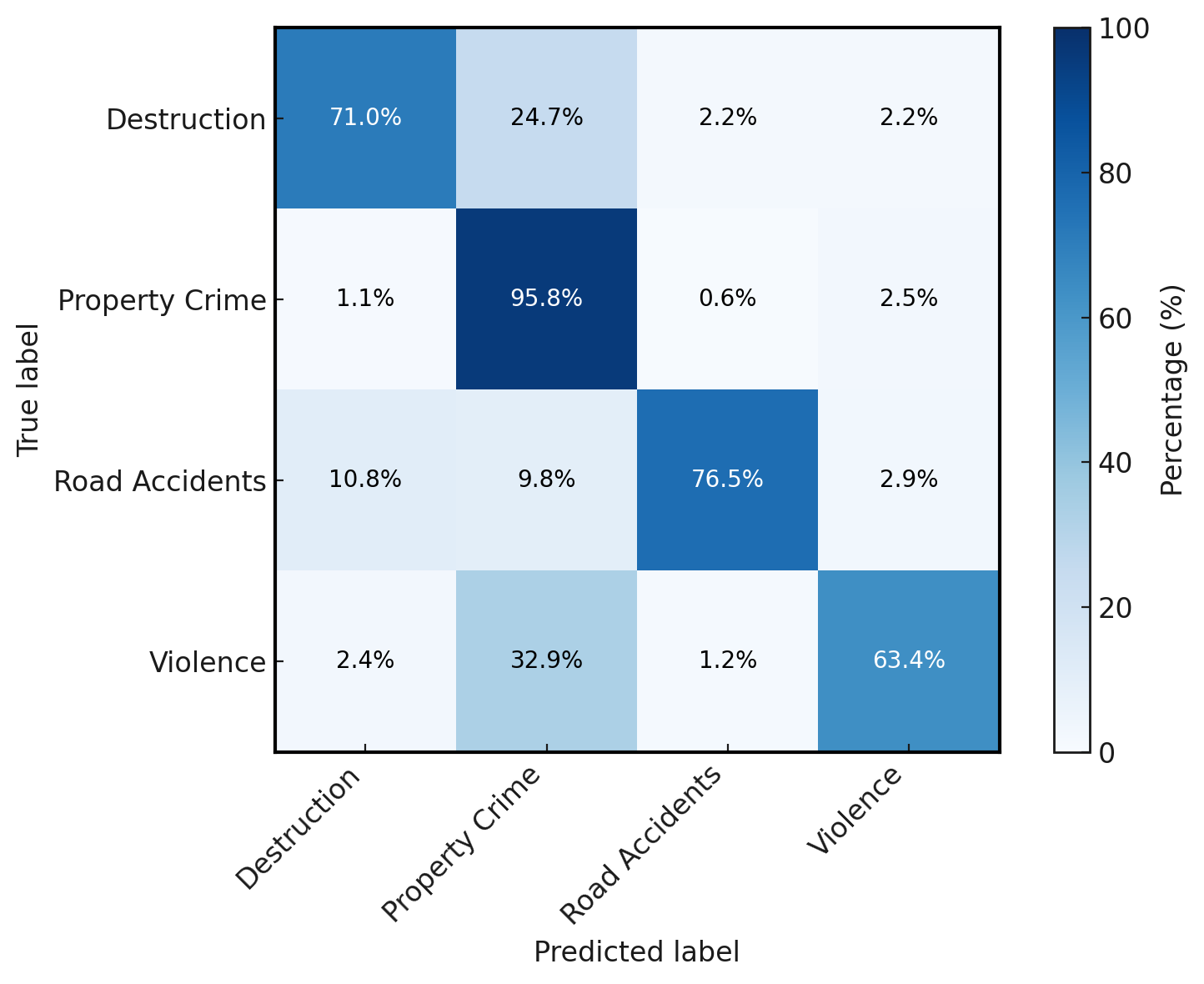}}
\caption{Normalized confusion matrix after semantic grouping of UCF-Crime classes into four categories. While overall classification performance improves, some residual confusion persists, especially between \textit{Violence} and \textit{Property Crime}.}
\label{fig:cm3}
\end{figure}

Figure~\ref{fig:cm3} shows the confusion matrix for \textbf{Ovis-8B}'s category inference. Overall misclassification rates are reduced relative to the previous matrix, except within the \textit{Property Crime} category, which still overlaps with \textit{Destruction} and \textit{Violence}.

\section{Conclusion}

We present a comprehensive head-to-head comparison of federated LoRA-tuned vision--language models (VLMs) and personalized lightweight 3D CNNs for video violence detection under realistic non-IID settings. All approaches exceed 90\% binary accuracy; the CNN achieves stronger calibration (ROC AUC: 92.59\%, log loss: 0.546) while using roughly half the energy of federated LoRA (240\,Wh vs.\ 570\,Wh). For multiclass, hierarchical grouping improves VLM accuracy (65.31\%~$\rightarrow$~81\%), and LoRA-based adaptation enhances VLM calibration under non-IID clients. Note that the CNN baseline was evaluated only for binary detection, whereas multiclass analysis was conducted exclusively with VLMs.

These findings crystallize an efficiency--capability trade-off: lightweight CNNs suit continuous, low-cost screening on constrained nodes, while VLMs should be selectively invoked for high-context reasoning and natural-language incident descriptions. We therefore recommend a hybrid cascade with always-on CNN screening and on-demand VLM escalation for complex or ambiguous events.

Our environmental accounting—via direct energy metering and emissions estimation—indicates that such tiered operation reduces energy and CO$_2$e without degrading situational awareness. Future work focuses on template-driven VLM reporting, uncertainty/novelty-conditioned model selection, multiclass 3D CNN, tighter PEFT, and lifecycle-aware assessment within \textbf{DIVA}.

\end{document}